\documentclass{article}

    \usepackage[preprint]{neurips_2020}

\usepackage{natbib}

\usepackage[utf8]{inputenc} % allow utf-8 input
\usepackage[T1]{fontenc}    % use 8-bit T1 fonts
\usepackage{hyperref}       % hyperlinks
\usepackage{url}            % simple URL typesetting
\usepackage{booktabs}       % professional-quality tables
\usepackage{amsfonts}       % blackboard math symbols
\usepackage{nicefrac}       % compact symbols for 1/2, etc.
\usepackage{microtype}      % microtypography
\usepackage{graphicx}
\usepackage{float}
\usepackage{array}
\usepackage{cite}
\usepackage{amsmath,amssymb}

\DeclareMathOperator*{\argmax}{arg\,max}

\DeclareMathOperator*{\hp}{\boldsymbol{\lambda}}
\DeclareMathOperator*{\HP}{\boldsymbol{\Lambda}}

\DeclareMathOperator*{\C}{\mathcal{C}}
\DeclareMathOperator*{\I}{I}

\title{Quantity vs. Quality: On Hyperparameter Optimization for Deep Reinforcement Learning}

\author{%
  Lars Hertel\\
  Department of Statistics\\
  University of California, Irvine\\
  Irvine, CA, 92697\\
  \texttt{lhertel@uci.edu} \\
   \And
   Pierre Baldi \\
   Department of Computer Science \\
University of California, Irvine \\
Irvine, CA 92697 \\
   \texttt{pfbaldi@uci.edu} \\
   \AND
  Daniel L. Gillen\\
  Department of Statistics\\
  University of California, Irvine\\
  Irvine, CA, 92697\\
  \texttt{dgillen@uci.edu} \\
}
\hypersetup{draft}
\begin{document}

\maketitle
\begin{abstract}
Reinforcement learning algorithms can show strong variation in performance between training runs with different random seeds. In this paper we explore how this affects hyperparameter optimization when the goal is to find hyperparameter settings that perform well across random seeds.
In particular, we benchmark whether it is better to explore a large \textit{quantity} of hyperparameter settings via pruning of bad performers, or if it is better to aim for \textit{quality} of collected results by using repetitions.
For this we consider the Successive Halving, Random Search, and Bayesian Optimization algorithms, the latter two with and without repetitions. We apply these to tuning the PPO2 algorithm \citep{schulman2017proximal} on the Cartpole balancing task and the Inverted Pendulum Swing-up task.
We demonstrate that pruning may negatively affect the optimization and that repeated sampling does not help in finding hyperparameter settings that perform better across random seeds. From our experiments we conclude that Bayesian optimization with a noise robust acquisition function is the best choice for hyperparameter optimization in reinforcement learning tasks.
\end{abstract}

% make the title area
\maketitle

\section{Introduction}
Many reinforcement learning methods can show significant variation in performance between runs with different random seeds \citep{henderson2018deep, islam2017reproducibility, nagarajan2018impact, colas2018many}.
This poses an issue for conducting hyperparameter optimization where the goal is to find hyperparameters configurations that perform well across random seeds.

Common hyperparameter optimization strategies typically assume exact observations or a negligible amount of uncertainty. For example, most hyperparameter optimization algorithms check different settings and then suggest the best observed. However, if observations vary drastically between random seeds, the best observed configuration may not be the configuration that performs best when averaged across many random seeds. Most model-free hyperparameter optimization methods such as random search \citep{bergstra2012random}, successive halving \citep{jamieson2016non, li2016hyperband}, and population-based training \citep{jaderberg2017population} do not explicitly handle uncertainty in observations. Evolutionary algorithms can in fact handle uncertainty \citep{loshchilov2016cma}. However, these methods may require more observations than is often feasible for hyperparameter optimization. Finally, model-based methods such as Bayesian optimization have a principled way to model uncertainty. However, they 
depend on an acquisition function to decide on the next query point \citep{picheny2013benchmark}. Not all acquisition functions are robust to uncertainty in the observations and especially the most commonly expected improvement is not. 

% While research on noisy Bayesian optimization exists \citep{letham2019constrained, wu2016parallel}, no clear evaluation has been done on the usefulness of these methods for noisy hyperparameter optimization. In particular, there are three important ways in which noisy hyperparameter search may differ from optimization on noisy synthetic benchmarks. First, objective functions in hyperparameter search are different and may be simpler than common optimization benchmarks. For example, the performance of an algorithm often marginally depends on a hyperparameter in a U-type way \citep{Goodfellow-et-al-2016}. Second, the noise may be heteroscedastic. In particular, there may be less noise near the optimum. Third, the noise distribution, that is how performance varies between random seeds, may be non-Gaussian. In particular, the noise may have limited support, may be skewed or may be close to uniform.

In order to deal with uncertainty in hyperparameter optimization, some researchers have adopted the practice of averaging across random seeds even during the hyperparameter search itself \citep{balandat2019botorch, falkner2018bohb, duan2016benchmarking}. Other sources \citep{stable-baselines} use hyperparameter optimization algorithms that are known to generally perform well but do not handle uncertainty. In this paper we assume that the goal is to find a hyperparameter optimization that performs well across random seeds. With that in mind, we benchmark different hyperparameter optimization strategies tuning a deep reinforcement learning algorithm on two continuous control tasks. %Section~\ref{sec:methods} briefly formalizes the optimization problem and then gives an overview of the benchmarked algorithms.

\section{Methods\label{sec:methods}}

\subsection{Problem Statement}
In order to define the problem we consider hyperparameters $\hp \in \HP$. We define the mean reward resulting from training with $\hp$ and a randomly picked seed as $\Psi(\hp)$. Finally, we define the expected mean reward across all random seeds for that hyperparameter settings as $\mathbb{E}[\Psi(\hp)]=f(\hp)$. The goal of the hyperparameter optimization here is to find 
\begin{equation}
    {\hp}^* = \argmax_{\hp \in \HP} f(\hp),
    \label{eq:noisy-optimization}
\end{equation}
that is the hyperparameter setting ${\hp}^*$ that yields the largest mean reward across random seeds.
The optimization in equation~\ref{eq:noisy-optimization} is difficult because $f(\hp)$ is unavailable and observations $\Psi(\hp)$ are computationally expensive. In particular, hyperparameter optimization faces two problems:
\begin{enumerate}
    \item Choosing where to evaluate the next ${\hp}_i$ during the optimization may be impacted by uncertainty, because previous observations are not exact.
    \item Choosing ${\hp}^*$ at the end of the optimization may be impacted, because the true performance of each ${\hp}_i$ is uncertain.
\end{enumerate}

\subsection{Repeated Evaluation\label{sec:repetitions}}
A simple way to approximate $f({\hp}_i)$ is to collect repeated evaluations and average the measurements $\bar{\Psi}({\hp}_i)=\sum_{j=1}^K \Psi({\hp}_i^j)$. While this approach is conceptually easy, the number of repetitions is often constrained by a computational budget. For this reason, the question is whether repetition is the optimal strategy or if it would be better to just evaluate more hyperparameter settings ${\hp}_i$. We now review a number of commonly used approaches to hyperparameter optimization and how these approaches are impacted by uncertainty.

\subsection{Model-free Optimization}
\subsubsection*{Random Search}
Random search is a simple and popular model-free hyperparameter search algorithm \citep{bergstra2012random}. In particular, random search can often serve as a simple but robust baseline against more complex methods \citep{li2019random}. In random search hyperparameter configurations $\hp$ are sampled uniformly at random from the specified hyperparameter space $\HP$. At the end of the optimization, the best observed hyperparameter setting ${\hp}^*$ is picked.

When applying random search in a setting where there is uncertainty, the suggestion of hyperparameter configurations is not impacted. However, the choice of the final recommended configuration may be influenced. In the presence of uncertainty the best observed setting may not be the on-average best setting. This issue can be alleviated using repetition as described in Section~\ref{sec:repetitions}, however at a higher computational cost. In experiments we denote random search as \textit{Random Search}, \textit{Random Search x3}, and \textit{Random Search x5} for using one, three, or five evaluations, respectively.
 
%One solution to this is to repeat evaluations of configurations. For example, instead of evaluating a configuration once, one can evaluate it five times and take the average of the validation errors. This reduces the noise, but may be computationally inefficient since a lot of computation can be spent on unpromising configurations.

% Comparison methods: SHA, RS with repetitions, SHA over repetitions
\subsubsection*{Bandit Algorithms}
Another more recent approach to hyperparameter search is the bandit-based \textit{Successive Halving Algorithm} (SHA) \citep{jamieson2016non}, as well as its extensions \textit{Hyperband} \citep{li2016hyperband}, and \textit{Asynchronous Successive Halving} (ASHA) \citep{li2018massively}. SHA and Hyperband have been shown to outperform random search and state-of-the-art model-based search methods on many hyperparameter optimization benchmarks from the domain of supervised learning.
SHA operates by randomly sampling configurations and allocating increasingly larger budgets to well-performing configurations. In the beginning, a small budget is allocated to each configuration. All configurations are evaluated and the top $1/\eta$ of configurations are promoted. Then the budget for each promoted configuration is increased by a factor of $\eta$. This is repeated until the maximum per-configuration budget is reached.
When tuning neural networks with SHA, the budget is usually the number of training iterations. That way the algorithm can probe a lot of configurations without fully evaluating them.
An algorithm that closely resembles SHA is used to tune hyperparameters of deep reinforcement learning algorithms in the Stable-baselines \citep{stable-baselines} package. We imitate the setup there by using the number of training steps as the resource. We refer to this as \textit{ASHA} in experiments since we are using the asynchronous version of SHA.

\subsubsection*{Evolutionary Algorithms}
Lastly, evolutionary algorithms have recently been successfully applied to hyperparameter optimization and architecture search and hyperparameter optimization for neural networks \citep{real2017large,loshchilov2016cma, hansen2019pycma}. Evolutionary algorithms such as CMA-ES \citep{hansen2003reducing} can handle high-dimensional search spaces and actually have a principled way of handling uncertainty. Nevertheless, evolutionary algorithms tend to require a larger number of function evaluations that is out of the scope of this study and may also be infeasible for many practitioners. For this reason, we leave the comparison to evolutionary algorithms for future work.

One algorithm from this paradigm that is popular with practitioners is Population-Based Training (PBT) \citep{jaderberg2017population}. In PBT a population of models is trained jointly with optimizing their hyperparameters. %This happens in a two-stage approach where all population members are first trained for a number of training iterations. In the second stage the population evolves by replacing bad performing members with good ones and perturbing their hyperparameters. The two stages alternate until all population members are fully trained.
Contrary to most other hyperparameter optimization approaches, PBT finds schedules of hyperparameters via perturbation and replacement between generations. PBT has shown strong results in multiple domains including the training of reinforcement learning agents.
The crucial difference in the goal of PBT and the goal we are aiming for in this paper is that PBT delivers one optimally trained agent. In this paper we however aim to obtain an optimal hyperparameter setting. While it is possible to trace back the schedule of hyperparameters of the best agent in PBT, we believe that it is less common to share and reproduce these. In addition, due to the fact that PBT finds schedules instead of specific hyperparameter settings it is difficult to accurately compare it to standard hyperparameter optimization methods which find fixed hyperparameters. For these reasons we do not include PBT in the experiments in Section~\ref{sec:experiments}.

% There is however a distinct difference in the goal of PBT compared to the other approaches discussed in this section. While most hyperparameter optimization methods aim to find the best performing hyperparameter setting, PBT aims to deliver an optimally trained agent. Retraining that model is complicated because a schedule of hyperparameter settings needs to be considered. For this reason we do not include PBT in our empirical evaluations.

\subsection{Bayesian Optimization}
\citep{bergstra2011algorithms} and \citep{snoek2012practical} demonstrated the feasibility of using Bayesian optimization for hyperparameter optimization. Bayesian optimization (BO) is a model-based approach to global derivative-free optimization of blackbox functions. Given existing observations of configurations and associated objective values, BO fits a surrogate model from the search space to the objective. This model is usually faster to evaluate than the objective function itself. For this reason, a large number of predictions of objective values can be made. An \textit{acquisition function} describes the utility of each prediction. By maximizing the acquisition function one can find the optimal next point to evaluate.

As described by \citep{srinivas2009gaussian}, it is common to use a Gaussian Process (GP) as a surrogate model for $f(\hp)$. For the covariance kernel the Matern 5/2 kernel is a common choice. Assuming normal observations noise, $\Psi(\hp) \sim \mathcal{N}(f(\hp), \tau^2)$ this offers a natural framework for uncertainty in observations. Inference for the observation variance $\tau^2$ can be made via maximization of the marginal likelihood. In particular, given a collection of observations $\C=\{(\hp_i, \Psi_i)\}_{i=1}^n$ and a GP prior, one can obtain the posterior distribution at hyperparameter setting $\hp$ as $f(\hp)|\C \sim \mathcal{N}(\mu_n(\hp), \sigma_n^2(\hp))$. Assuming a zero prior, the posterior mean and variance are given by:
\begin{align*}
    \mu_n(\hp)&=\mathbf{k}(\hp)^T(\mathbf{K} + \tau^2 \mathbf{I})^{-1} \\
    \sigma^2_n(\hp) &= k(\hp, \hp) -  \mathbf{k}(\hp)^T (\mathbf{K} + \tau^2 \mathbf{I})^{-1} \mathbf{k}(\hp)
\end{align*}
where $\mathbf{k}(\hp)$ is the vector of covariances with all previous observations and $\mathbf{K}$ is the covariance matrix of the observations.

\subsubsection*{Expected Improvement}
As mentioned before, Bayesian optimization chooses the next point to evaluate via the acquisition function. The most popular choice of acquisition function for hyperparameter optimization is the expected improvement (EI) \citep{mockus1978application}. The EI is given by
\begin{equation}
    EI(\hp)=\mathbb{E}_{f(\hp)|\C}[\I( \Psi_{min}, f(\hp))].
\end{equation}
where $\I(r, s) = \max(0, r - s)$ and
$\Psi_{min} = \min_{1 \leq i \leq n} \Psi_i$ is the minimum observed validation loss. Part of why the EI is popular is that it can be computed analytically. However, if observations are uncertain $\mathrm{Var}(\Psi(\hp))$ may be large and $\Psi_{min}$ may just be low by chance. This can result in over-exploitation around $\Psi_{min}$ and the optimization getting "stuck" around this point. In experiments we refer to EI as \textit{GPyOpt-EI} since we are using the GPyOpt \citep{gpyopt2016} implementation. We also show experiments for \textit{GPyOpt-EI x3}, referring to EI with averaging across three repetitions for each hyperparameter setting.

% Go over alternatives
% Expected Improvement with Plug-in
% An obvious alternative is then to want to use $\min_{\hp} \mu(\hp)$ instead of $\Psi_{min}$. This approach is also referred to as \textit{plug-in EI} by \citep{picheny2013benchmark}.
% % AEI
% \citep{huang2006global} extend this idea via the Augmented Expected Improvement (AEI). In this case, the expected improvement is defined as
% \begin{equation}
%     AEI(\hp)=\mathbb{E}[\max(0, T-\mu(\hp)]\left(1-\frac{\tau}{\sigma^2(\hp)+\tau^2}\right).
% \end{equation}
% Here, $\tau^2$ is the residual variance as estimated by the GP, $T=\mu(\hp^{**})$, and $\hp^{**}=\argmin_{\hp} \mu(\hp)+\alpha \sigma(\hp)$. The constant $\alpha$ can be picked by the user or left to the default setting of 1 recommended by \citep{huang2006global}. \citep{picheny2013benchmark} found the AEI to be empirically competitive for noisy optimization.

% NEI
\subsubsection*{Noisy Expected Improvement}
\citep{letham2019constrained} extended expected improvement to handle uncertainty in observations by taking the expectation over observations. Ignoring constraint handling also introduced by the authors, the noisy expected improvement (NEI) can be written as:
% \begin{equation}
%     NEI(\hp) = \int_{\mathbf{f}_n} EI(\hp|\mathbf{f}_n)p(\mathbf{f}_n|\Psi_1,\ldots,\Psi_n)d\mathbf{f}_n
% \end{equation}
% \begin{equation}
%     NEI(\hp) = \mathbb{E}_{f(\hp_{1:n})|\C} \left[ \mathbb{E}_{f(\hp)|\C^*}[\max(f(\hp) - \max_{\hp_i \in \hp_{\mathrm{obs}}} f(\hp_i), 0)] \right].
% \end{equation}
\begin{equation}
\label{eq:NEI}
    NEI(\hp) = \mathbb{E}_{f(\hp_{1:n})|\C} \{ \mathbb{E}_{f(\hp)|\mathcal{C^*}}[\I( \Psi_{min}^*, f(\hp))] \}.
\end{equation}
where $\mathcal{C^*}=\{(\hp_i, \Psi^*(\hp_i)):\Psi^*(\hp_i)\sim f(\hp_i)|\mathcal{C}\}_{i=1}^n$ is a fantasy dataset sampled from the posterior and $\Psi_{min}^*$ is the minimum of the fantasy dataset.
In practice, this approach fits a GP model. GP samples are drawn from the posterior at the observed points. These samples represent fantasy datasets. Each fantasy dataset is treated as in the regular EI, that is, a GP is fitted and the EI is computed across the space $\HP$. The EI values are then averaged across the fantasy datasets to obtain the NEI. \citet{balandat2019botorch} simplify this approach by taking the expectation over the joint posterior:
\begin{equation}
\label{eq:qNEI}
    qNEI(\hp) = \mathbb{E}_{f(\hp),f(\hp_{1:n})|\C} [\I(\min f(\boldsymbol{\lambda}_{1:n}),f(\hp)].
\end{equation}
We note that Equation~\ref{eq:NEI} and~\ref{eq:qNEI} are both computationally intensive and non-trivial to implement. However, \citet{letham2019constrained} showed that NEI outperforms regular EI on objective functions with Gaussian noise and given the true variance. In our experiments we use the qNEI implemented in BoTorch\footnote{\url{https://botorch.org/}} to calculate the noisy expected improvement. Contrary to the experiments in \citep{letham2019constrained} we infer the variance $\tau^2$. We refer to this approach as \textit{BoTorch-qNEI}.% We further note that the true distribution of $\Psi(\hp)$ may be non-gaussian which could impact the performance of NEI. %We also perform experiments with the mean of repeated evaluations in which case a noise estimate will be provided and the distribution of the mean will be closer to normal due to the central limit theorem. %The qNEI with repeated samples we will call \textit{BoTorch-qNEI x3}.
% \begin{equation}
%     qNEI(\hp) = \mathbb{E}[\max(\xi) - \max(\xi_{\mathrm{obs}})]
% \end{equation}
% where $(\xi, \xi_{\mathrm{obs}}) \sim N([\mu(\hp),\mu(\hp_{\mathrm{obs}})], \Sigma_{[\hp, \hp_{\mathrm{obs}}]})$.
% Explain NEI in detail

% LCB
\subsubsection*{Lower Confidence Bound}
An alternative to the EI criterion is the Lower(/Upper) Confidence Bound (LCB) criterion \citep{srinivas2009gaussian}. The LCB takes the form
\begin{equation}
    LCB({\hp}) = \mu_n(\hp) + \beta \sigma_n(\hp)
\end{equation}
where $\beta$ is a coefficient set by the user. This coefficient is sometimes termed the \textit{exploration weight}. Tuning the exploration weight may improve the performance of LCB. However, since in practice the computational budget to tune this parameter is rarely available, we use a default value of $\beta=2$.
While LCB has been shown to be outperformed by EI on noise-less objective functions \citep{snoek2012practical}, contrary to EI it is not impacted by uncertainty in the observations. This is due to the fact that it only depends on the posterior mean and variance. In all experiments we refer to LCB as \textit{GPyOpt-LCB}.

We note that there are more available acquisition functions than the ones presented in this section. In particular, the knowledge gradient \citep{scott2011correlated, wu2016parallel} is an acquisition function that, while difficult to compute, can handle uncertainty and has been shown to outperform the presented acquisition functions \citep{balandat2019botorch}. Furthermore, entropy-search-based methods \citep{hennig2012entropy, hernandez2014predictive, wang2017max} also naturally handle uncertainty. We leave the evaluation of these methods to future work.

% this is all about picking the next point
While the choice of acquisition function determines how to choose the next point, one still needs to pick a best setting at the end of the optimization. In the noise-less case this is often done by picking the best observed hyperparameter $\hp_i$. \citet{frazier2018tutorial} suggested to rely on the GP model and pick the best predicted $\hp$. Since this approach is robust to uncertainty we use the predicted best $\hp$ in all experiments. To make this feasible in multi-dimensional tasks we find the best predicted by optimizing an LCB acquisition with $\beta=0$, thereby utilizing existing optimization tools in BoTorch and GPyOpt.

% now regarding the choosing the best point
\section{Experiments\label{sec:experiments}}

% \subsection{Evaluation}
% Explain how we obtain xbest_pred and y_best pred
% For the evaluation of all methods we consider, where applicable, the predicted best observation at the end of the optimization as suggested by \citep{frazier2018tutorial}. This is opposed to the usual more conservative approach of picking the best observed setting. In particular, we find the $\hp$ value that corresponds to the lowest predicted objective value. To compare the different methods we then consider the true expected objective value at that $\hp$. For model-free approaches such as random search and SHA we will use the best observed $\hp$.

\subsection{Cartpole Balancing\label{sec:cartpole}}
\begin{figure}[H]
    \centering
    \fbox{\includegraphics[width=\linewidth]{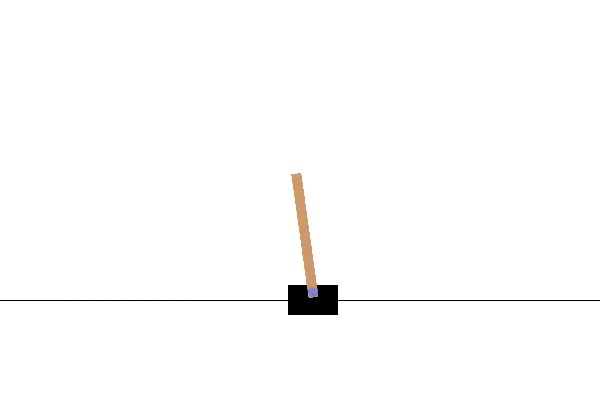}}
    \caption{Screenshot of the cartpole task.}
    \label{fig:cartpole-screenshot}
\end{figure}

First, we compare the introduced algorithms on the cartpole balancing task. The cartpole task \citep{barto1983neuronlike} is a continuous control problem that requires to balance a pole on a friction-less cart. The cart can be moved left or right. A screenshot of the task is shown in Figure~\ref{fig:cartpole-screenshot}.

The cartpole task has been used as a benchmark for reinforcement learning (RL) algorithms \citep{duan2016benchmarking} and most recently to benchmark hyperparameter optimization for reinforcement learning \citep{falkner2018bohb, balandat2019botorch}. We tune the hyperparameters of the proximal policy optimization algorithm by \citet{schulman2017proximal}. For this we use the implementation from Stable-baselines \citep{stable-baselines} and the Cartpole-v1 environment from OpenAI Gym \citep{brockman2016openai}. In particular, we optimize the log-learning-rate (uniform $[-5,-1]$) as well as the entropy-coefficient (uniform $[0,1]$), discount-factor (uniform $[0,1]$), and likelihood-ratio-clipping (uniform $[0,1]$).
% The RL agent is trained for 30000 steps. RL agents on the cartpole task are often compared in terms of the number of episodes until a mean reward of at least 195 is obtained over a one hundred episode run. However, this measure is not optimal for hyperparameter optimization since some agents may not converge. Alternatively, RL agents are often compared via plotted reward curves. The qualitative comparison of two reward curves can be made quantitative by considering the area under the reward curve. For a constant number of steps the area under the curve is proportional to the mean reward across the training. We thus choose the maximization of the mean reward across training as the hyperparameter optimization objective.
The RL agent is trained for 30000 steps and we use the mean reward across training as the objective for the hyperparameter optimization.

\begin{figure*}
\centering
\includegraphics[width=\linewidth]{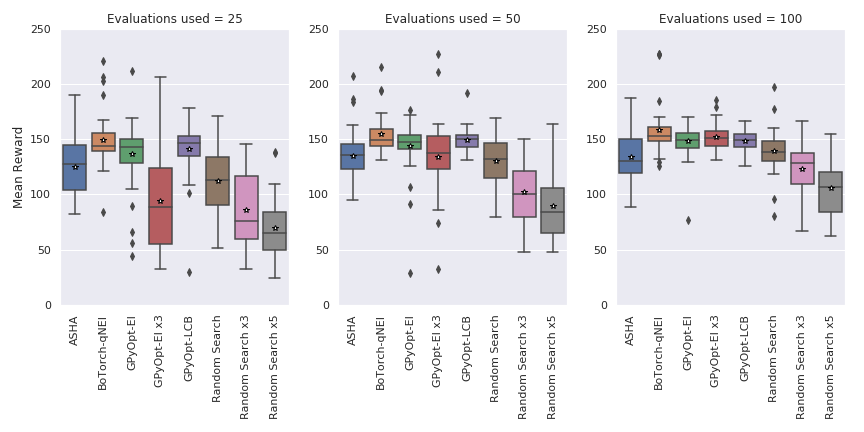}
\caption{Boxplots showing the distribution of mean rewards across training over 40 optimization runs on the Cartpole environment using PPO2 (higher is better).\label{fig:cartpole-results}}
\end{figure*}

The hyperparameter optimization itself is allowed to train 100 agents. This translates to 100 hyperparameter settings for methods without repetition and for example 20 hyperparameter settings if five repetitions per setting were used. After the budget is used up, the best configuration is trained 20 times with different random seeds and the mean across those runs is taken as the hyperparameter optimization outcome. Figure~\ref{fig:cartpole-results} (right) shows results for 40 hyperparameter optimization runs conducted in this fashion. Evaluation is also done after 25 and 50 agents have been trained (Figure~\ref{fig:cartpole-results} left and middle, respectively). Corresponding means are shown Table~\ref{tab:cartpole-table}.

\begin{table}[]
\caption{Cartpole balancing mean rewards across training and hyperparameter optimization runs for different budgets. Cell values are averaged across 20 training runs of the best found agent and across 40 hyperparameter optimization runs.}
\label{tab:cartpole-table}
\centering
\begin{tabular}{lrrr}
\toprule
Evaluations &    25 &    50 &   100 \\
\midrule
ASHA             & 125.2 & 135.0 & 134.4 \\
BoTorch-qNEI     & \textbf{149.7} & \textbf{155.2} & \textbf{158.3} \\
GPyOpt-EI        & 136.5 & 143.6 & 148.5 \\
GPyOpt-EI x3     &  93.7 & 133.5 & 151.9 \\
GPyOpt-LCB       & 141.1 & 149.7 & 148.7 \\
Random Search    & 112.0 & 130.3 & 139.1 \\
Random Search x3 &  85.6 & 102.6 & 123.0 \\
Random Search x5 &  69.6 &  89.1 & 105.7 \\
\bottomrule
\end{tabular}
\end{table}

From Figure~\ref{fig:cartpole-results} and Table~\ref{tab:cartpole-table} it can be seen that \textit{BoTorch-qNEI} has the best results across all budgets, however the difference to other Bayesian optimization methods is small. \textit{GPyOpt-EI x3} performs worse in the beginning, but later catches up. This suggests that its three repetitions do not help in this case and would be better used to explore additional hyperparameter settings. \textit{ASHA} seems to have some very good runs, but stagnates at a slightly worse performance than the Bayesian optimization methods. \textit{Random Search} performs almost on-par with the Bayesian optimization methods after 100 evaluations. \textit{Random Search x3} and \textit{Random Search x5} perform worst overall, but do show steady improvement. In this case more evaluations may be needed for these methods to perform better.

% it is notable that \textit{GPyOpt-LCB} is only marginally worse. \textit{GPyOpt-EI} catches up after 50 evaluations and \textit{GPyOpt-EI x3} after 100 evaluations. This seems to indicate that the EI acquisition does perform worse due to noise. For EI with 3 replicates it seems that the computation allocated to replicates does not help, especially on a budget of 25 evaluations. One would expect that \textit{Random Search} eventually catches up with the model-based methods. This is however not the case here. This may be due to the fact that \textit{Random Search} still needs more evaluations. It may also be because \textit{Random Search} picks the best observed configuration which may be inaccurate. \textit{Random Search x3} and \textit{Random Search x5} seem to lack a sufficient number of observations. Finally, ASHA, here applied to early stopping of configurations seems to stagnate after 50 evaluations. We believe that the pruning of configurations after partial training may be the cause of this. Finally, we also note that \textit{BoTorch-qNEI} has some outliers with significantly better mean rewards. It seems therefore that there is still room for improvement on the benchmarked algorithms.

\subsection{Inverted Pendulum Swing-up}
\begin{figure}[H]
    \centering
    \fbox{\includegraphics[width=0.7\linewidth]{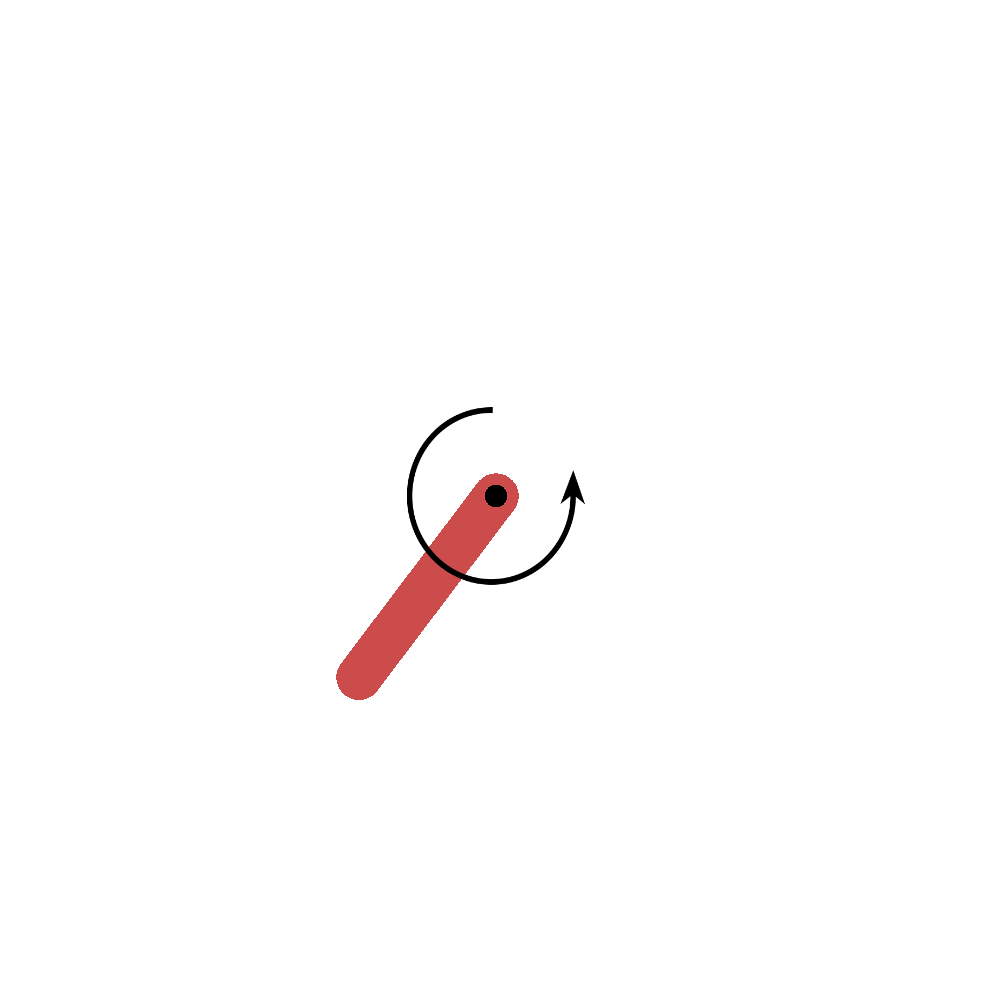}}
    \caption{Screenshot of the Inverted Pendulum Swing-up task.}
    \label{fig:pendulum-screenshot}
\end{figure}

As a second experiment, we consider the Inverted Pendulum Swing-up task. Just like the Cartpole balancing task from Section~\ref{sec:cartpole}, the Inverted Pendulum Swing-up task is a classic control problem. A screenshot of the environment is shown in Figure~\ref{fig:pendulum-screenshot}. The friction-less pendulum (red) starts in a random position with random velocity and the goal is to keep it standing up. To achieve this the agent can apply negative or positive effort to the joint. The reward is a function of how upright and still the pendulum is.

In comparison to the Cartpole balancing task, this problem is considered more difficult. We again apply the proximal policy algorithm by \citep{schulman2017proximal}. This time the hyperparameter search-space consists of the log-learning-rate (uniform $[-5,-1]$) and the log-entropy-coefficient (uniform $[-5,0]$). The agent is given $2\times 10^6$ timesteps for training. The hyperparameter optimization objective is, as before, the mean reward across training.

\begin{figure*}
\centering
\includegraphics[width=\linewidth]{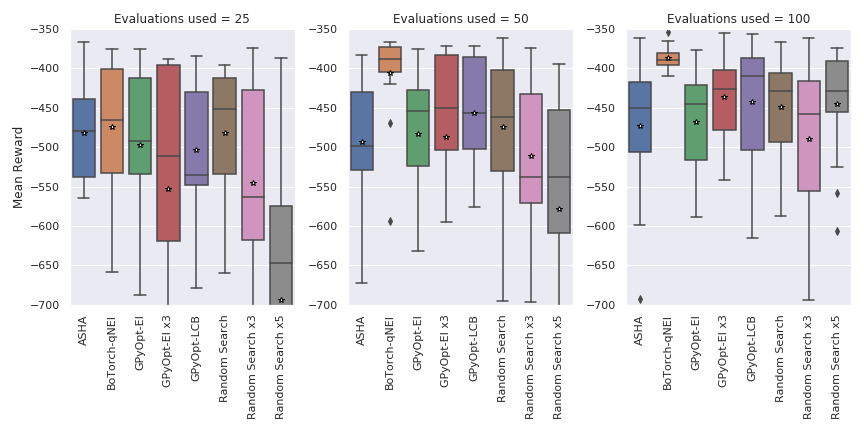}
\caption{Boxplots showing the distribution of mean rewards across training over 15 hyperparameter optimization runs on the Inverted Pendulum Swing-up task using PPO2 (higher is better).\label{fig:pendulum-results}}
\end{figure*}

Figure~\ref{fig:pendulum-results} shows boxplots of mean rewards across training over 15 hyperparameter optimization runs. The hyperparameter optimization is again allowed to train 100 agents. Additional evaluations are again done after 25 and 50 trained agents. It can be seen that after 25 evaluations all algorithms perform similarly except for \textit{Random Search x5} which performs worse. After 50 evaluations \textit{BoTorch-qNEI} clearly performs better than all other algorithms. This trend continues after 100 evaluations and except for \textit{BoTorch-qNEI} no clear difference can be seen in the performance of the different algorithms. In terms of the median \textit{GPyOpt-LCB} still performs second-best, however the distribution of its outcomes is very overlapping with that of the other methods. Table~\ref{tab:pendulum-table} shows corresponding means, nevertheless we believe that it is important to consider the distribution of outcomes for each method and believe Figure~\ref{fig:pendulum-results} better represents the results of this study.
\begin{table}[]
\caption{Mean rewards for the Inverted Pendulum Swing-up task across training and across hyperparameter optimization runs for different budgets (higher cell values are better). Cell values are averaged across 20 training runs of the best found agent and across 15 hyperparameter optimization runs.}
\label{tab:pendulum-table}
\centering
\begin{tabular}{lrrr}
\toprule
Evaluations &    25 &    50 &   100 \\
\midrule
ASHA             & -482.0 & -493.7 & -472.6 \\
BoTorch-qNEI     & \textbf{-474.7} & \textbf{-406.2} & \textbf{-387.3} \\
GPyOpt-EI        & -497.2 & -483.5 & -467.5 \\
GPyOpt-EI x3     & -553.1 & -487.6 & -436.9 \\
GPyOpt-LCB       & -503.7 & -456.2 & -442.4 \\
Random Search    & -481.8 & -474.2 & -449.3 \\
Random Search x3 & -546.1 & -511.4 & -490.2 \\
Random Search x5 & -694.0 & -578.2 & -444.6 \\
\bottomrule
\end{tabular}
\end{table}

\section{Discussion}
We evaluated different hyperparameter optimization methods on two reinforcement learning benchmarks with the goal of finding hyperparameter settings that perform well across random seeds. The recorded outcome of each hyperparameter optimization is therefore the mean reward of the best setting averaged across multiple random seeds. From comparing algorithms on two reinforcement learning tasks we have found that model-based search in the form of GP-based Bayesian optimization performs best in terms of the average achieved across optimization runs. While in the Cartpole experiment all Bayesian optimization methods perform similarly, in the Inverted Pendulum the noise-robust expected improvement clearly outperforms other methods.

% We believe that this is due to the fact that the GP estimates the mean function whose minimum is our optimization target. Among Bayesian optimization methods we find that the lower confidence bound as well as noisy expected improvement perform best. This is intuitive since these acquisition functions are able to handle noise better than the expected improvement acquisition function.

We find that, contrary to experiments from supervised learning, evaluating more configurations by using pruning as in the Successive Halving algorithm does not outperform Bayesian optimization in this case.

Furthermore, using replicates for each hyperparameter setting, as often done for reinforcement learning, does not outperform regular methods. For example, random search with one sample outperforms random search with three or five repetitions in both experiments when computation is matched. Using replication for Bayesian optimization does also not seem to improve performance.

% using replicates does not necessarily provide better results, both for bayesian optimization and random search. Overall, our conclusion is that the best strategy for noisy hyperparameter optimization is to use a model-based search. The lower confidence bound performed well in experiments, is easy to implement, and is available in most Bayesian optimization packages. If feasible, more advanced criteria such as the noisy expected improvement may provide further improvements.

% Focussed on continuous parameters

% Future work on this topic includes the evaluation of Bayesian optimization methods that are not GP based. In addition to that further improvements are possible as shown in the Cartpole results of Section~\ref{sec:cartpole}. Making noisy expected improvement more robust to non-Gaussian noise may be one direction towards such improvements.
\section{Future Work}
Future work on this topic include the evaluation of model-based multi-fidelity optimization such as introduced by \citep{poloczek2017multi} or \citep{wu2019practical}. The question is whether such methods are better able to delineate noise even for partial evaluations of hyperparameter settings. Another direction for future work is to better model the observations of the objective, specifically with regards to the noise distribution. As mentioned before, NEI depends on the assumption of Gaussian noise and its performance may deteriorate if the observation noise is significantly non-Gaussian.

\section*{References}

\bibliographystyle{plainnat}
\bibliography{main}

\end{document}